\documentclass{article}

\usepackage[final,nonatbib]{neurips_2019_ml4ps}

\usepackage[utf8]{inputenc} 
\usepackage[T1]{fontenc}    
\usepackage{hyperref}       
\usepackage{url}            
\usepackage{booktabs}       
\usepackage{amsfonts}       
\usepackage{nicefrac}       
\usepackage{microtype}      
\usepackage{algorithm2e}    

\usepackage{graphicx}
\usepackage{xcolor}
\usepackage{wrapfig}         
\usepackage{floatrow}

\title{Emulation of physical processes with Emukit}

\author{%
  Andrei Paleyes \\
  University of Cambridge\thanks{Work done while all authors were at Amazon Research Cambridge.} \\
  \texttt{ap2169@cam.ac.uk} \\
  \And
  Mark Pullin \\
  Arrival\footnotemark[1] \\
  \texttt{pullin@arrival.com} \\
  \And
  Maren Mahsereci \\
  University of T\"ubingen\footnotemark[1] \\
  \texttt{maren.mahsereci@uni-tuebingen.de} \\
  \And
  Cliff McCollum \\
  Facebook\footnotemark[1] \\
  \texttt{cmccollum@fb.com} \\
  \And
  Neil Lawrence \\
  University of Cambridge\footnotemark[1] \\
  \texttt{ndl21@cam.ac.uk} \\
  \And
  Javier Gonzalez \\
  Microsoft Research Cambridge\footnotemark[1] \\
  \texttt{Gonzalez.Javier@microsoft.com}
}

\newcommand{\seir}{\textsc{seir}}

\begin{document}

\maketitle

\begin{abstract}
Decision making in uncertain scenarios is an ubiquitous challenge in real world systems. Tools to deal with this challenge include simulations to gather information and statistical emulation to quantify uncertainty.
The machine learning community has developed a number of methods to facilitate decision making, but so far they are scattered in multiple different toolkits, and generally rely on a fixed backend. 
In this paper, we present \emph{Emukit}, a highly adaptable Python toolkit for enriching decision making under uncertainty. 
Emukit allows users to:
(i) use state of the art methods including Bayesian optimization, multi-fidelity emulation, experimental design, Bayesian quadrature and sensitivity analysis; 
(ii) easily prototype new decision making methods for new problems. Emukit is agnostic to the underlying modeling framework and enables users to use their own custom models. 
We show how Emukit can be used on three exemplary case studies.

\end{abstract}

\section{Intro: ML for physical processes and simulations}
\label{sec:intr-ml-simul}
Machine learning has demonstrated a huge success in several problems in the \emph{virtual world}. 
Today, complex data representations can be learned using neural networks, enabling state of the art predictions in problems like image classification, natural language processing, reinforcement learning, or recommender systems.
 When applying machine learning in the \emph{physical world} however, one faces different challenges. 
A key aspect of problems in robotics or climate modeling is in building models able to encapsulate physical knowledge and data.
Computer simulators are widely used in scientific research and industry to analyse the behaviour of those complex systems \cite{OHagan98uncertaintyanalysis}. 
These simulations map some knowledge about a system of interest into code that is able to reproduce real-world scenarios under varying conditions. 
Different types of simulators exist, ranging from computational fluid dynamics to optimization solvers. 
Simulators can also be used to test a system before it is built, reducing the burden and sometimes impractical limitations of real experimentation. 
In this paper, we define a simulator as any computer program that imitates a real-world system or process.

Despite the clear advantages of using simulations, they also come with three main drawbacks: (i)~a simulator can only simulate what it has been programmed to simulate, which can bias the analysis of how the system will work in the real world; (ii)~a simulator can be slow and expensive to run, which is an issue when its running time exceeds the time frame needed for making a decision or when they need to be used at scale; (iii)~sensitivity and uncertainty analyses are impractical for expensive simulators, as thousands of simulation runs are typically required to perform these tasks.
All three issues can be addressed by \emph{statistical emulation}, also called surrogate modelling, where the basic idea is to replace the simulator (or parts of it) with machine learning models --- emulators, that use inputs and outputs of a system to make quick predictions about new and previously unobserved calls. 
For this, a few simulations are run and used to train a machine learning model that can predict the outputs of the simulator given the inputs. 
Common choices of statistical emulators are Gaussian processes (GPs), \cite{Rasmussen2006}, random forests \cite{Ho:1995:RDF:844379.844681} or Bayesian neural networks \cite{MacKay:1992:PBF:148147.148165}.
An emulator is therefore a ‘model of a model’. 
It is a statistical model of the simulator, which is itself a mechanistic model of the world. 
Interestingly, simulators are often based on deterministic models while emulators, which are probabilistic by nature, enable quantifying uncertainty that is crucial for principled decision making. 
There are different types of uncertainty an emulator may attempt to quantify, especially when combining data from simulations and real world: the uncertainty that the emulator has about the simulator, the uncertainty the simulator has about the real world as well as uncertainty that comes from numerical approximations we make when running simulations (see e.g., \cite{HenOsbGir15}).

Some seminal references about emulation are \cite{kennedy1999bayesian} and \cite{conti2009, conti2007bayesian}. In modern systems design emulators are used in robotics \cite{Peherstorfer2017SurveyOM}, aircraft design \cite{sun2014}, control \cite{DeisenrothRT2011} as well as to understand extremely complex processes such as the climate \cite{Stefano2014}. A key aspect that makes them useful is that they allow us to reason probabilistically about \emph{outer-loop decisions} such as optimization \cite{DBLP:journals/pieee/ShahriariSWAF16} and data collection \cite{Krause2007NonmyopicAL}, and to use them to explain how uncertainty propagates in a system \cite{Bilionis2016}. Interestingly, it is also possible to build emulators that combine real and simulated data. These emulators are called multi-fidelity methods and have become increasingly relevant during the last few years \cite{Peherstorfer2017SurveyOM}. When properly combined with outer-loop applications, multi-fidelity emulation is a powerful tool for efficient systems design via transfer learning \cite{marco_ICRA_2017}. 
These methods correct biases in simulations by including real-world data. 
Building emulators for low-level numerical methods has also captured the attention of the scientific community leading to the emerging field of probabilistic numerics \cite{HenOsbGir15}.

In the past years, the machine learning community has developed powerful software tools that link statistical emulation and decision making. Some examples are GPyOpt \cite{gpyopt2016}, GPFlowOpt \cite{GPflowOpt2017} and BoTorch\footnote{https://www.botorch.org/} libraries for Bayesian optimization or Bayesquad\footnote{https://github.com/OxfordML/bayesquad} for Bayesian quadrature. 
Offering off-the-shelf methods, these frameworks are generally conveniently build on top of a hardcoded model backend, e.g. GPyOpt and Bayesquad are relying on GPy \cite{gpy2014} and numpy, GPFlowOpt on Tensorflow \cite{tensorflow2015whitepaper} and BoTorch on PyTorch \cite{paszke2017automatic}. This dependency is practical as it enables black-box algorithms and fast software development. On the flip side, it is impossible or extremely tedious to switch out the model backend for a custom user model, or a code component for a prototype, e.g., a novel sampler, optimizer, surrogate model etc. Switching out components is essential for fair comparison and benchmarking, especially when frameworks are used in a scientific context.
Another key observation is that existing frameworks generally tackle only one out of the set of decision problems at a time. This separation seems artificial as many of the decision solvers exhibit a common loop structure (see Section~\ref{sec:emuk-softw-pack}) and it is wasteful not to re-use that common code.
Alternatively one may argue that existing frameworks do not naturally lend themselves to using the same emulator model in several decision problems which may lead to inconsistent models and decisions when used in a common task or application. 


To address these issues our paper contains the following contributions:

\begin{itemize}
\item We introduce \emph{Emukit}, highly adaptable Python toolkit for enriching decision making under uncertainty. Emukit, to our knowledge, is unique because it is agnostic to the model backend, and combines several tools for emulation of physical or mechanical processes, uncertainty quantification and decision making, under one umbrella.
\item We discuss several cases where Emukit was used, to highlight its flexibility and user friendliness.
\end{itemize}
Section~\ref{sec:emuk-softw-pack} introduces the Emukit software package and its unique features, Section~\ref{sec:probl-solv-workfl} discusses a typical workflow with Emukit, and Section~\ref{sec:showcases} shows three case studies to illustrate typical tasks for which Emukit may be used.


\section{Emukit software package}
\label{sec:emuk-softw-pack}
Bayesian optimization, Bayesian experimental design and Bayesian quadrature are all decision making processes that follow a similar pattern. 
Algorithmically, they can be thought of as instances of a common abstract loop.
As a software package, Emukit is structured in the same way (Algorithm~\ref{alg:emukit_loop}).

Emukit is built in a way in which all components of Algorithm~\ref{alg:emukit_loop} interact through loosely coupled interfaces. This ensures modularity, enabling researchers to easily replace a particular element (for example, use a different point collection strategy) without affecting the rest of the decision making loop.
\begin{algorithm}[t]
	\caption{Decision making loop in the Emukit toolkit}
	\label{alg:emukit_loop}
    \While{stopping condition is not met}{
        select next system inputs by some optimal decision\;
        gather observations by running the system of interest with selected inputs\;
        update emulator model with new observation(s)\;
    }
\end{algorithm}
A notable feature of Emukit that separates it from other packages that provide similar decision making functionality is its independence from modeling backends. Users of Emukit toolkit can choose any existing Python scientific computing engine (numpy, Tensorflow, etc.), any modeling library on top of this engine, implement a few Emukit interfaces, and then use the rest of the library without any changes. This way we encourage code re-use and fair comparison in the community. 

Some of the features that make Emukit a unique tool are:
\begin{itemize}
\item \emph{Connect to custom models:} Emukit separates models from decisions. Methods like Bayesian optimization, multi-arm bandits, experimental design (active learning) or Bayesian quadrature are decision processes that depend on some surrogate model of the system of interest. Emukit allows users to train such models and apply them in several decision scenarios.
\item \emph{Agnostic to model backend:} Emukit is agnostic to the underlying modelling framework. This means users can utilize any tool of their choice in the Python ecosystem to build the machine learning model and apply it in a decision loop. Models written in Scipy, GPy, TensorFlow, MXnet, etc. can easily be wrapped in Emukit.
\item \emph{Modular and extendable:} Emukit has been built using reusable components. Most Emukit components (optimizers, samplers, etc.) can be used in isolation. When connected via a series of clear APIs they can be used to build complex decisions tools.
This modularity eases the creation of new probabilistic methods for uncertainty quantification and decision making. Benchmarking by means of replacing individual components thus becomes a natural application of the library.
\end{itemize}

Emukit has a lightweight installation process: it can be installed by just running \textit{pip install emukit}.

\subsection{Workflow with Emukit}
\label{sec:probl-solv-workfl}
\begin{figure}[h]
  \centering
    \includegraphics[scale=0.4]{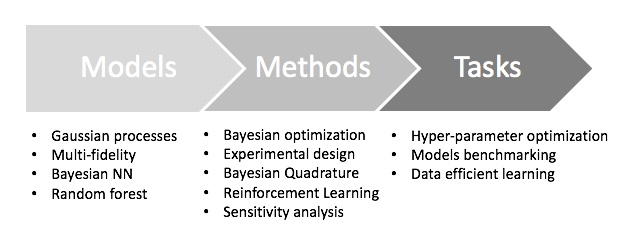}  
  \caption{Summary of workflow for the users of Emukit. A model is computed in modelling framework of choice. The model is wrapped using a pre-defined interface and connected to the core components of several methods such as Bayesian optimization, experimental design etc. Specific tasks are then solve using these methods.}
  \label{fig:workflow}
\end{figure}

The typical workflow for users of Emukit is as follows (see Figure~\ref{fig:workflow} for a graphical description):
\begin{itemize}
\item \emph{Build the model:} Instead of constraining the user to certain model classes, Emukit provides the flexibility of using user-specified models. Generally speaking, Emukit does not provide modelling capabilities, instead, expecting users to define their own models. Because of the variety of modelling frameworks available, Emukit does not mandate or make any assumptions about a particular modelling technique or a library. Instead, it suggests to implement a subset of defined model interfaces that are required to use a particular method.
\item \emph{Run the method:} This is the main focus of Emukit. Emukit defines a general structure of a decision making method, called OuterLoop (essentially an implementation of Algorithm~\ref{alg:emukit_loop}), and then offers implementations of several such methods: Bayesian optimization, Bayesian quadrature, experimental design, and sensitivity analysis. All methods in Emukit toolkit are model-agnostic and defining new APIs to accommodate other frameworks is easy.
\item \emph{Solve the task:} For the end users, Emukit is a way to solve a certain task, which may have research or business value. Emukit comes with a set of examples of how tasks such as hyper-parameter tuning, sensitivity analysis multi-fidelity modelling or benchmarking are accomplished using the library.
\end{itemize}
\section{Showcases}
\label{sec:showcases}
The following three subsection discuss case studies where Emukit has been utilized successfully. 
They illustrate how Emukit toolkit can be used in practice. 
Further practical examples can be found in the Emukit codebase.

\subsection{Estimate the expected infection peak of the SEIR epidemic model}
\label{sec:sir-model}
The authors of \cite{Gessner19} used Emukit to compute expected outcomes of a black-box simulation that models the spread of an infectious disease. 
The research questions the authors answered were: what are the expected \emph{time occurrence}, and \emph{height} (in number of individuals) of the infection peak, given that the infection rate is uncertain? Both answers require the computation of an intractable, one-dimensional integral, where evaluating the integrand equals running the simulation once (ground truth integrals take approximately 1 hour on a laptop each). 
The authors used the Bayesian quadrature loop in Emukit to solve both integrals efficiently.

The simulation is based on the \seir~epidemic model \cite{Kermack1927} whose dynamics are governed by stochastic discrete-time events of individuals changing infection state (see \cite{Gessner19} or \cite{Daley1999} for details).
The so-called Gillespie algorithm samples trajectories from the process, hence, to query the \seir~model, many thousand runs are required which can then be averaged to construct estimators.  
Due to its sample efficiency, Emukit's Bayesian quadrature loop completes in a few minutes only.
Acquiring both solutions merely require the user to wrap each simulation output as an Emukit user-function. Alternatively, if a custom backend is preferred, the user could hand a Gaussian process model over to Emukit, and Emukit would integrate the model automatically. 
Emukit's emulator models as well as the inferred distributions of the corresponding estimators of the height and time of the infection peak are shown in Figure~\ref{fig:seir} in the appendix.

\subsection{Optimizing the design of a superconducting resonator}
\label{sec:quant-comp-stuff}

Quantum computers promise to revolutionize computer science \cite{Calude17}. 
The largest quantum computers built to date make use of superconducting qubit architectures. 
These architectures have many advantages but suffer from low decoherence times, where the decoherence of qubits means the wave function collapses and the information stored in the qubits is lost.
To overcome this limitation, information can be transferred to a more robust storage medium where the information is encoded in particle spins. 
To transfer information from qubits to spins, superconducting resonators are used \cite{Majer07}. 
This case study describes how a superconducting resonator can be designed using Emukit for this purpose. 
It is being conducted by a research group in UCL, and is currently in progress, therefore we are only able to describe the process for which Emukit is used.

Bayesian optimization is used to optimize the design of a superconducting resonator for a quantum memory. 
A particular design is parameterized by its most important geometric parameters as shown in Figure~\ref{fig:resonator} in the appendix. 
Three separate figure of merits were defined that describe different desirable physical properties of the resonator. 
Emukit is then used to optimize a weighted sum of the figure of merits with respect to the geometric properties of the resonator. 
Each evaluation of the objective function involves running a numerical simulation that computes the electromagnetic properties of the resonator and a subsequent calculation of the figures of merit.


\subsection{Reverse engineering a musical synthesizer}
\label{sec:synthesizer-example}
A musical synthesizer produces sound by generating waveforms via oscillators.
Created audio streams are then routed through a pipeline that consists (not necessary all) of mixing of separate streams, filtering, adding of noise, and saturation (see Figure~\ref{fig:synth} in appendix, left). 
Musicians can control the output sound by changing the configuration of the pipeline. 
This gives rise to the following common problem: given some target sound, find a synthesizer configuration that outputs this target \cite{Garcia_growingsound}.

The authors of \cite{pmlr-v89-uhrenholt19a} decided to solve this problem with Bayesian optimization. 
In order to estimate the discrepancy between the produced sound and the target, they designed a novel modeling approach, in which Gaussian process is used to model the distribution of the output's L2 norm. 
The flexibility of Emukit allowed to implement this customization directly, without necessary effort duplication. 
Emukit's API also provided functionality to perform the task of a fair comparison to standard Bayesian optimization. The results of this comparison can be seen on Figure~\ref{fig:synth}, right (appendix).

\section{Conclusion}
\label{sec:disc--concl}
We presented Emukit, a highly adaptable Python toolkit for enriching decision making under uncertainty. 
We discussed favorable features of Emukit such as its modularly, model backend independence, extensibility, and easy benchmarking. 
We showed a range of completed and ongoing projects where Emukit was utilized as a tool. 
In the future we are planning to continue expanding the usage of Emukit to new domains and applications. We are also hoping to generalize the idea of Emukit beyond Bayesian methods to other areas of active learning.


\bibliographystyle{unsrt}
\bibliography{bibfile}

\begin{thebibliography}{10}

\bibitem{OHagan98uncertaintyanalysis}
A.~O'Hagan, J.~M. Bernardo, J.~O. Berger, A.~P. Dawid, A.~F. M.~Smith (eds,
  M.~C. Kennedy, and J.~E. Oakley.
\newblock Uncertainty analysis and other inference tools for complex computer
  codes.
\newblock 1998.

\bibitem{Rasmussen2006}
C.~E. Rasmussen and C.~Williams.
\newblock Gaussian processes for machine learning.
\newblock 2006.

\bibitem{Ho:1995:RDF:844379.844681}
T.~K. Ho.
\newblock Random decision forests.
\newblock In {\em Proceedings of the Third International Conference on Document
  Analysis and Recognition (Volume 1) - Volume 1}, ICDAR '95, pages 278--,
  Washington, DC, USA, 1995. IEEE Computer Society.

\bibitem{MacKay:1992:PBF:148147.148165}
D.~J.~C. MacKay.
\newblock A practical bayesian framework for backpropagation networks.
\newblock {\em Neural Comput.}, 4(3):448--472, May 1992.

\bibitem{HenOsbGir15}
P.~Hennig, M.~A. Osborne, and M.~Girolami.
\newblock Probabilistic numerics and uncertainty in computations.
\newblock {\em Proceedings of the Royal Society of London A: Mathematical,
  Physical and Engineering Sciences}, 471(2179), 2015.

\bibitem{kennedy1999bayesian}
M.C. Kennedy and A.~O'Hagan.
\newblock {\em Bayesian Calibration of Computer Models}.
\newblock University of Sheffield, Department of Probability and Statistics,
  1999.

\bibitem{conti2009}
S.~Conti, J.~P. Gosling, J.~E. Oakley, and A.~O'Hagan.
\newblock Gaussian process emulation of dynamic computer codes.
\newblock {\em Biometrika}, 96(3):663--676, 2009.

\bibitem{conti2007bayesian}
S.~Conti and A.~O'Hagan.
\newblock {\em Bayesian Emulation of Complex Multi-output and Dynamic Computer
  Models}.
\newblock Research report (University of Sheffield. Department of Probability
  and Statistics). Department of Probability \& Statistics, University of
  Sheffield, 2007.

\bibitem{Peherstorfer2017SurveyOM}
B.~Peherstorfer and K.~Willcox.
\newblock Survey of multifidelity methods in uncertainty propagation,
  inference, and optimization.
\newblock 2017.

\bibitem{sun2014}
Y.~G. Sun and L.~Sun.
\newblock The design of avionics system interfaces emulation and verification
  platform based on qar data.
\newblock In {\em Mechanical Components and Control Engineering III}, volume
  668 of {\em Applied Mechanics and Materials}, pages 879--883. Trans Tech
  Publications, 11 2014.

\bibitem{DeisenrothRT2011}
M.~P. Deisenroth and C.~E. Rasmussen.
\newblock Pilco: A model-based and data-efficient approach to policy search.
\newblock In {\em Proceedings of the 28th International Conference on Machine
  Learning, ICML 2011}, pages 465--472. Omnipress, 2011.

\bibitem{Stefano2014}
S.~Castruccio, D.~J. McInerney, M.~L. Stein, F.~L. Crouch, R.~L. Jacob, and
  E.~J. Moyer.
\newblock Statistical emulation of climate model projections based on
  precomputed gcm runs.
\newblock {\em Journal of Climate}, 27(5):1829--1844, 2014.

\bibitem{DBLP:journals/pieee/ShahriariSWAF16}
B.~Shahriari, K.~Swersky, Z.~Wang, R.~P. Adams, and N.~de~Freitas.
\newblock Taking the human out of the loop: {A} review of bayesian
  optimization.
\newblock {\em Proceedings of the {IEEE}}, 104(1):148--175, 2016.

\bibitem{Krause2007NonmyopicAL}
A.~Krause and C.~Guestrin.
\newblock Nonmyopic active learning of gaussian processes: an
  exploration-exploitation approach.
\newblock In {\em ICML}, 2007.

\bibitem{Bilionis2016}
I.~Bilionis and N.~Zabaras.
\newblock {\em Bayesian Uncertainty Propagation Using Gaussian Processes},
  pages 1--45.
\newblock Springer International Publishing, Cham, 2016.

\bibitem{marco_ICRA_2017}
A.~Marco, F.~Berkenkamp, P.~Hennig, A.~P. Schoellig, A.~Krause, S.~Schaal, and
  S.~Trimpe.
\newblock Virtual vs. {R}eal: Trading off simulations and physical experiments
  in reinforcement learning with {B}ayesian optimization.
\newblock In {\em Proceedings of the IEEE International Conference on Robotics
  and Automation (ICRA)}, pages 1557--1563, May 2017.

\bibitem{gpyopt2016}
The~GPyOpt authors.
\newblock {GPyOpt}: A bayesian optimization framework in python.
\newblock \url{http://github.com/SheffieldML/GPyOpt}, 2016.

\bibitem{GPflowOpt2017}
N.~Knudde, J.~{van der Herten}, T.~Dhaene, and I.~Couckuyt.
\newblock {{GP}flow{O}pt: {A} {B}ayesian {O}ptimization {L}ibrary using
  Tensor{F}low}.
\newblock {\em arXiv preprint -- arXiv:1711.03845}, 2017.

\bibitem{gpy2014}
{GPy}.
\newblock {GPy}: A gaussian process framework in python.
\newblock \url{http://github.com/SheffieldML/GPy}, since 2012.

\bibitem{tensorflow2015whitepaper}
M.~Abadi, A.~Agarwal, P.~Barham, E.~Brevdo, Z.~Chen, C.~Citro, G.~S. Corrado,
  A.~Davis, J.~Dean, M.~Devin, S.~Ghemawat, I.~Goodfellow, A.~Harp, G.~Irving,
  M.~Isard, Y.~Jia, R.~Jozefowicz, L.~Kaiser, M.~Kudlur, J.~Levenberg,
  D.~Man\'{e}, R.~Monga, S.~Moore, D.~Murray, C.~Olah, M.~Schuster, J.~Shlens,
  B.~Steiner, I.~Sutskever, K.~Talwar, P.~Tucker, V.~Vanhoucke, V.~Vasudevan,
  F.~Vi\'{e}gas, O.l Vinyals, P.~Warden, M.~Wattenberg, M.~Wicke, Y.~Yu, and
  X.~Zheng.
\newblock {TensorFlow}: Large-scale machine learning on heterogeneous systems,
  2015.
\newblock Software available from tensorflow.org.

\bibitem{paszke2017automatic}
A.~Paszke, S.~Gross, C.~Chintala, G.~Chanan, E.~Yang, Z.~DeVito, Z.~Lin,
  A.~Desmaison, L.~Antiga, and A.~Lerer.
\newblock Automatic differentiation in {PyTorch}.
\newblock In {\em NIPS Autodiff Workshop}, 2017.

\bibitem{Gessner19}
A.~Gessner, J.~Gonzalez, and M.~Mahsereci.
\newblock {A}ctive {M}ulti-{I}nformation {S}ource {B}ayesian {Q}uadrature.
\newblock In {\em Conference on Uncertainty in Artificial Intelligence}, 2019.

\bibitem{Kermack1927}
W.~O. Kermack and A.G. McKendrick.
\newblock A contribution to the mathematical theory of epidemics.
\newblock {\em {Proceedings of the Royal Society of London A: Mathematical,
  Physical and Engineering Sciences}}, 115(772):700--721, 1927.

\bibitem{Daley1999}
D.~J. Daley and J.~Gani.
\newblock {\em Epidemic Modelling: An Introduction}.
\newblock Cambridge Studies in Mathematical Biology. Cambridge University
  Press, 1999.

\bibitem{Calude17}
C.~S. Calude and E.~Calude.
\newblock The road to quantum computational supremacy, 2017.

\bibitem{Majer07}
J.~Majer, J.~M. Chow, J.~M. Gambetta, J.~Koch, B.~R. Johnson, J.~A. Schreier,
  L.~Frunzio, D.~I. Schuster, A.~A. Houck, A.~Wallraff, A.~Blais, M.~H.
  Devoret, S.~M. Girvin, and R.~J. Schoelkopf.
\newblock Coupling superconducting qubits via a cavity bus.
\newblock {\em Nature}, 449:443 EP --, 09 2007.

\bibitem{Garcia_growingsound}
R.~A. Garcia.
\newblock Growing sound synthesizers using evolutionary methods.
\newblock In {\em Proceedings of ALMMA 2002 Workshop on Artificial Models for
  Musical Applications}, pages 99--107, 2001.

\bibitem{pmlr-v89-uhrenholt19a}
A.K. Uhrenholt and B.S. Jensen.
\newblock Efficient bayesian optimization for target vector estimation.
\newblock In Kamalika Chaudhuri and Masashi Sugiyama, editors, {\em Proceedings
  of Machine Learning Research}, volume~89 of {\em Proceedings of Machine
  Learning Research}, pages 2661--2670. PMLR, 16--18 Apr 2019.

\end{thebibliography}

\newpage

\appendix
\section*{Appendix: Figures}

\begin{figure}[h]
  \centering
    \includegraphics[scale=0.2]{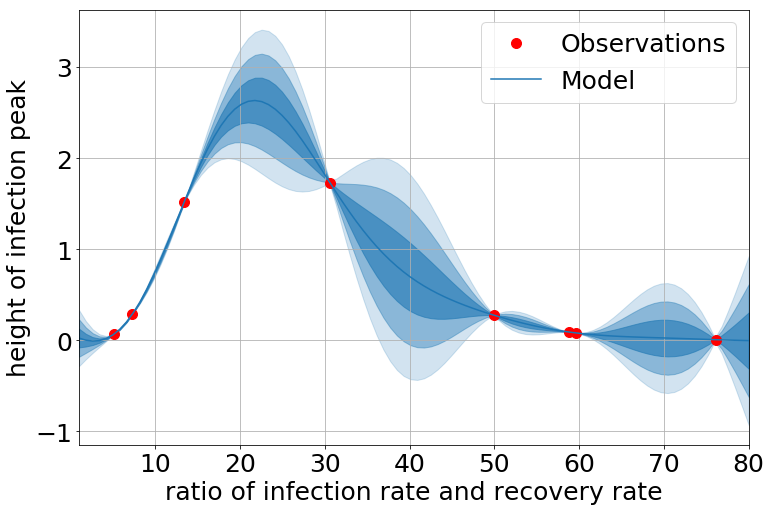}  
    \includegraphics[scale=0.2]{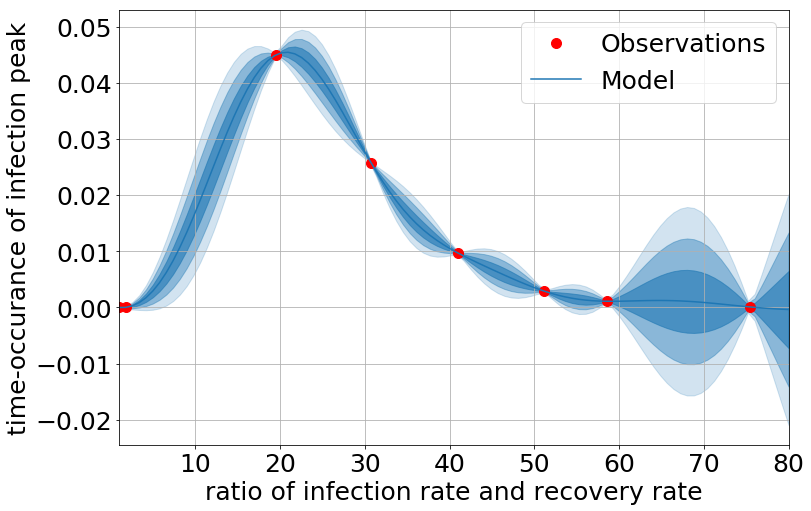}  \\
    \includegraphics[scale=0.2]{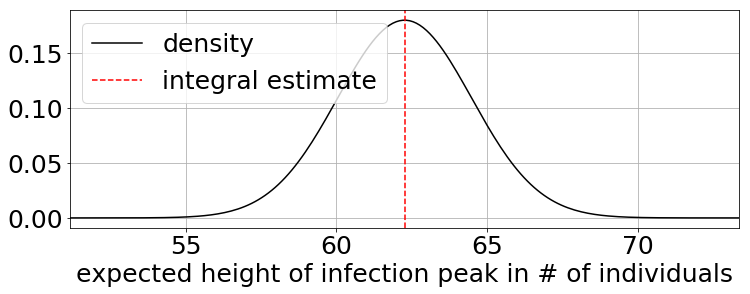}  
    \includegraphics[scale=0.2]{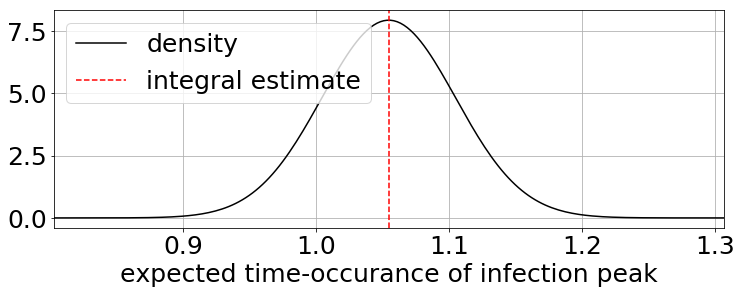}  
  \caption{Emukit for the numerical integration.
\emph{Left:} expected height of infection peak. \emph{Right:} expected time of infection peak. 
\emph{Top row:} emulator model of the integrands. 
\emph{Bottom row:} inferred distributions of the corresponding estimators. 
The estimated expected height and time of the infection peak is $62.0\pm 5$ individuals and $1.05\pm 0.1$ time units respectively (one credible interval).
}
  \label{fig:seir}
\end{figure}
\begin{figure}[h]
  \centering
    \includegraphics[height=4.5cm, width=6.5cm]{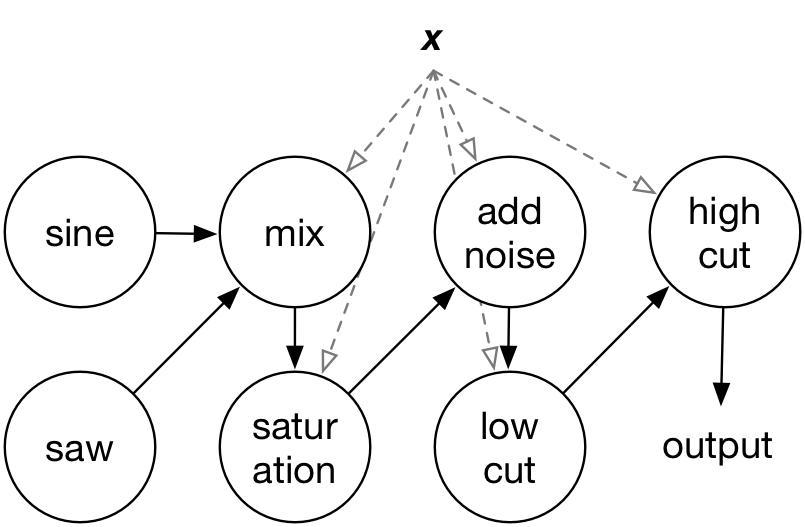}
    \includegraphics[height=4.5cm, width=6.5cm]{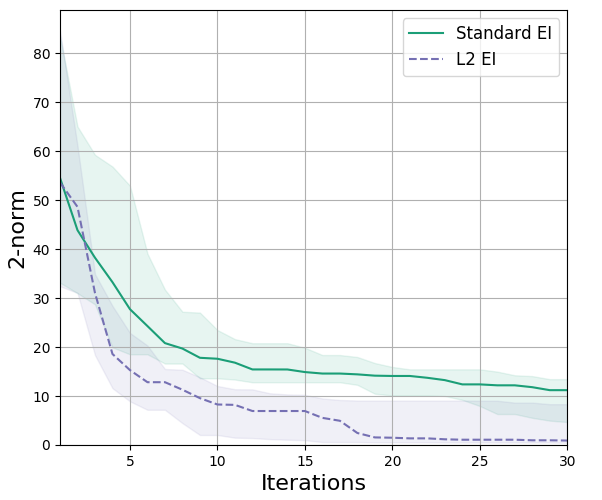}
  \caption{Reverse engineering a musical synthesizer. Left: synthesizer pipeline. Right: synthesizer's pipeline optimization towards a given sound, comparison between \cite{pmlr-v89-uhrenholt19a} and a standard Bayesian optimization.}
  \label{fig:synth}
\end{figure}
\begin{figure}[h]
\floatbox[{\capbeside\thisfloatsetup{capbesideposition={right,top},capbesidewidth=6cm}}]{figure}[\FBwidth]
{\caption{Basic design of the ART resonator. The red outline indicated the capacitive section of the design and the blue indicates the inductive section. The parameters which can be varied are: a)~capacitor length, b)~capacitor gap width, c)~capacitor width, d)~inductor length, e)~inductor width. Other parameters which aren't shown are: inductor gap width and height. Emukit is used to optimize physical properties of the resonator, which directly depend on these parameters.}
\label{fig:resonator}}
{\includegraphics[height=7cm, width=7cm]{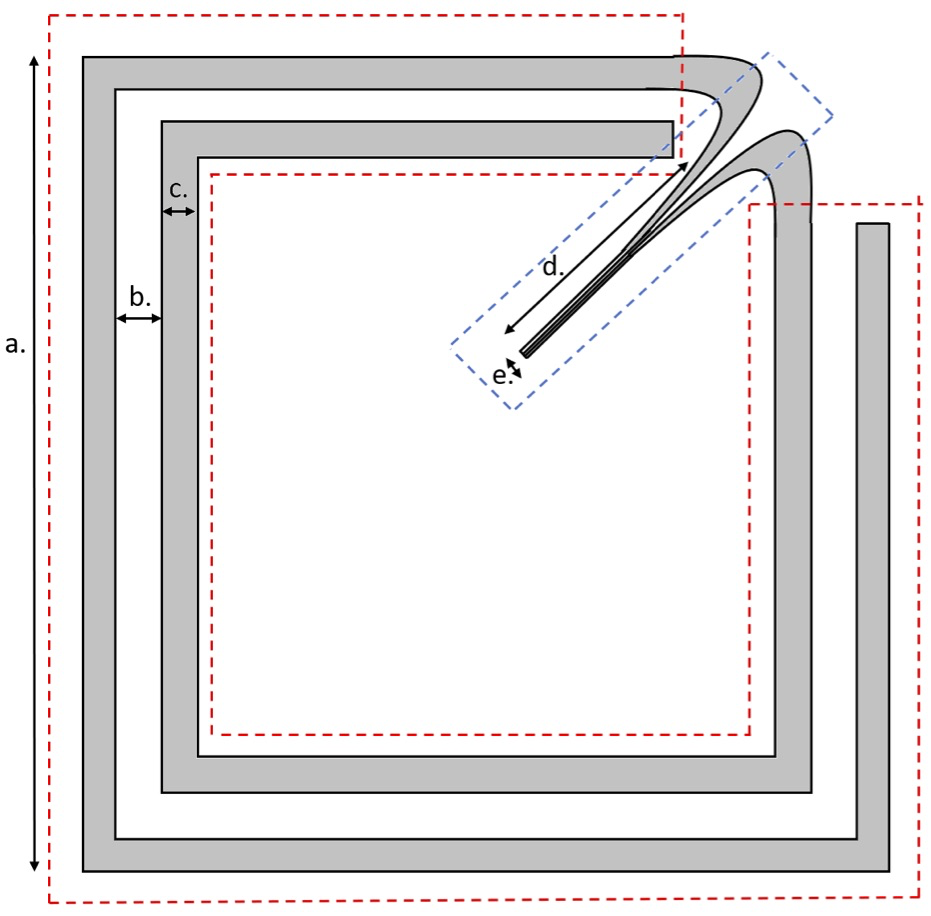}}
\end{figure}
\end{document}